\documentclass[conference]{IEEEtran}
\IEEEoverridecommandlockouts
\usepackage[T1]{fontenc}

\usepackage{cite}
\usepackage{amsmath,amssymb,amsfonts}
\usepackage{algorithmic}
\usepackage{graphicx}
\usepackage{textcomp}
\usepackage{url}

\usepackage{xcolor}
\def\BibTeX{{\rm B\kern-.05em{\sc i\kern-.025em b}\kern-.08em
    T\kern-.1667em\lower.7ex\hbox{E}\kern-.125emX}}
\begin{document}

\title{Embodied GPT-5.1: Evidence of a World Model?
\thanks{Research sponsored by PECE - Graduate Studies in Artificial Intelligence and by Coordenação de Aperfeiçoamento de Pessoal de Nível Superior - CAPES (PROAP 2050). Thiago Martins is Sponsored by CNPq (grant 309688/2025-6) }
}

\author{\IEEEauthorblockN{Roberto Spinelli}
\IEEEauthorblockA{\textit{Grad. Studies in Mechanical Eng.} \\
\textit{University of São Paulo}\\
São Paulo, Brazil \\
rspinelli@usp.br}
\and
\IEEEauthorblockN{Thiago C. Martins}
\IEEEauthorblockA{\textit{Department of Mechatronics} \\
\textit{University of São Paulo}\\
São Paulo, Brazil \\
thiago@usp.br}
}
\maketitle
\let\thefootnote\relax\footnote{\textcopyright{} 2026 IEEE. Personal use of this material is permitted. Permission from IEEE must be obtained for all other uses, in any current or future media, including reprinting/republishing this material for advertising or promotional purposes, creating new collective works, for resale or redistribution to servers or lists, or reuse of any copyrighted component of this work in other works. DOI: 10.1109/CROS69211.2026.11565684.}
\begin{abstract}
This exploratory study examines whether a large multimodal language model, GPT-5.1, can serve as the high-level controller of a physical mobile robot despite having no prior embodiment, no training in simulated environments, and no exposure to sensorimotor experience. Using only low-resolution first-person images and a discrete action set, the model was tasked with navigation and object-directed behaviors such as locating and contacting a target toy. Across multiple trials, GPT-5.1 demonstrated emergent capabilities that suggest elements of spatial reasoning and physical understanding. These included maintaining short-term memory of object locations after they left the camera frame, inferring the physical consequences of its own movements, and executing coherent action sequences such as colliding with an object and reversing to visually verify the outcome. At the same time, the model displayed inefficiencies and perceptual limitations, including imprecise alignment strategies and occasional misidentification of distant distractors. Overall, the results indicate that GPT-5.1 exhibits signs of world-model-like behavior in an embodied setting, despite the absence of any embodiment-related training, a finding that challenges long-standing views in cognitive science and robotics which hold that a physical body is a necessary prerequisite for developing such forms of intelligence. The findings motivate deeper investigation into the emergence, limits, and robustness of physical understanding in large language models.
\end{abstract}


\section{Embodied Intelligence and the Role of Physicality}

Theoretical frameworks in cognitive science and robotics converge on the view that intelligence is fundamentally embodied. Grounded cognition holds that cognitive processes depend on sensorimotor interaction, bodily states, and situated action~\cite{barsalou2008grounded, shapiro2019embodied}, while behavior-based robotics rejects internal symbolic models in favor of the principle that “the world is its own best model”~\cite{brooks1991intelligence}. Pfeifer and Scheier~\cite[Ch. 3]{pfeifer1999understanding} similarly argue that only physically situated agents can cope with real-world physics and grounding. Sensorimotor theories describe perception as mastery of sensorimotor contingencies linking action to sensory change~\cite{oregan2001sensorimotor}, and developmental robotics shows that learning such contingencies is essential for self-organization and cognitive growth~\cite{jacquey2019sensorimotor}. Harnad’s symbol grounding problem~\cite{harnad1990symbol} formalizes the concern that, without sensorimotor referents, symbolic systems lack intrinsic meaning.

There is significant recent progress in the integration of LLMs into robotics systems.
Hierarchical architectures such as PaLM-SayCan~\cite{ahn2022can} use LLMs for high-level planning while delegating execution to low-level controllers. 
Vision-language systems like PaLM-E~\cite{driess2023palm} and EmbodiedGPT~\cite{mu2023embodiedgpt} take first-person images and instructions as input to generate goal-directed action sequences, and systems such as LangNav~\cite{pan2024langnav} and RobotGPT~\cite{jin2024robotgpt} explore perception-to-text pipelines and LLM-generated policies.
Sikorski \textit{et al.}~\cite{Sikorski2025EdgeLLM} use general-purpose LLMs (GPT-4-Turbo and quantized LLaMA-2) on a desktop to interpret spoken commands and convert them into low-level motion primitives on a small Arduino-based wheeled robot.
Although those approaches have shown that cloud LLMs can reliably parse and sequence locomotion commands, they do not handle onboard first-person vision or embodied visual grounding, and their language models operate purely as text-to-action translators.
The authors further note that offline LLMs remain unreliable for physical control, underscoring the difficulty of grounding generic models in real-world execution.

Modern machine learning approaches to robotics control problems in complex environments have formalized the concept of the World Model (WM) as the learning of a compressed spatial and temporal representation of the environment~\cite{HaSchmidhuber2018}.
Drawing on cognitive analogies, the WM is an internal representation of the physical environment that enables the agent to generate accurate predictions about the consequences of its actions.
Contemporary research in Embodied AI (EAI) recognizes the core challenge of bridging high-level semantics with grounded physical interaction~\cite{10.1145/3746449}. Multimodal Large Language Models (MLLMs), such as GPT-5.1, excel at semantic reasoning and task decomposition but often overlook physical constraints and lack strong adaptation to environmental feedback. Conversely, traditional WMs provide physics-awareness and internal representations, yet struggle with semantic reasoning. Therefore, a critical research question remains: Can an embodied MLLM, which has not been explicitly trained on physical simulation, develop the emergent capabilities required for a functional world model?

\section{Objective}
In this exploratory study, we ask whether models trained exclusively on text and images—without any physical body, and without access to simulated or real environments during training—can develop the ability to act through a physical robot in the real world. Can such models support spatial orientation, navigation, and task-directed physical behavior? More broadly, can a disembodied model develop something akin to proprioceptive awareness when placed in control of a body for the first time?

These questions touch on deeper issues: do such interactions reveal the emergence of an internal World Model—an inner representation that encodes the structure of the physical environment and enables the model to generate accurate predictions about the consequences of its own actions? We aim to explore these questions by evaluating the ability of GPT-5.1 to act through a physical robot in the real world.

\section{System Architecture}
\begin{figure}
\centering
\includegraphics[width=0.8\columnwidth]{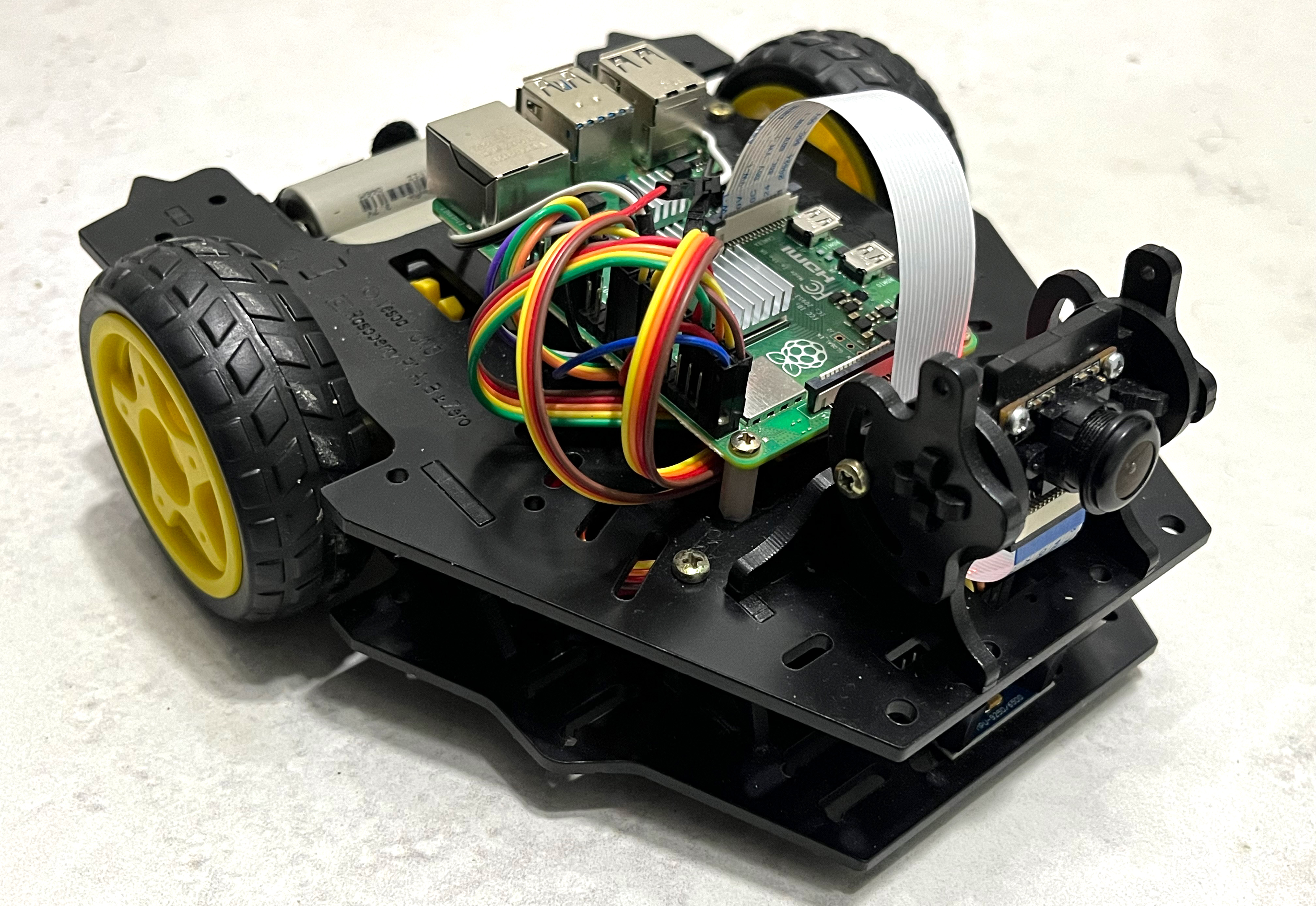}
\caption{The differential-drive mobile robot platform equipped with a forward-facing camera used in this study.}
\label{fig:robot_photo}
\end{figure}

This study uses a differential-drive mobile robot equipped with a forward-facing camera and basic odometry. The platform has been previously described in detail in~\cite{11066122}, including its mechanical structure, sensor suite, and embedded control electronics. For completeness, we summarize only its essential characteristics here: a two-wheel differential drive base, onboard battery supply, a microcontroller responsible for motor actuation, and a camera module providing real-time first-person visual input. Figure~\ref{fig:robot_photo} shows our little robot.

\begin{figure}
\centering
\includegraphics[width=0.8\columnwidth]{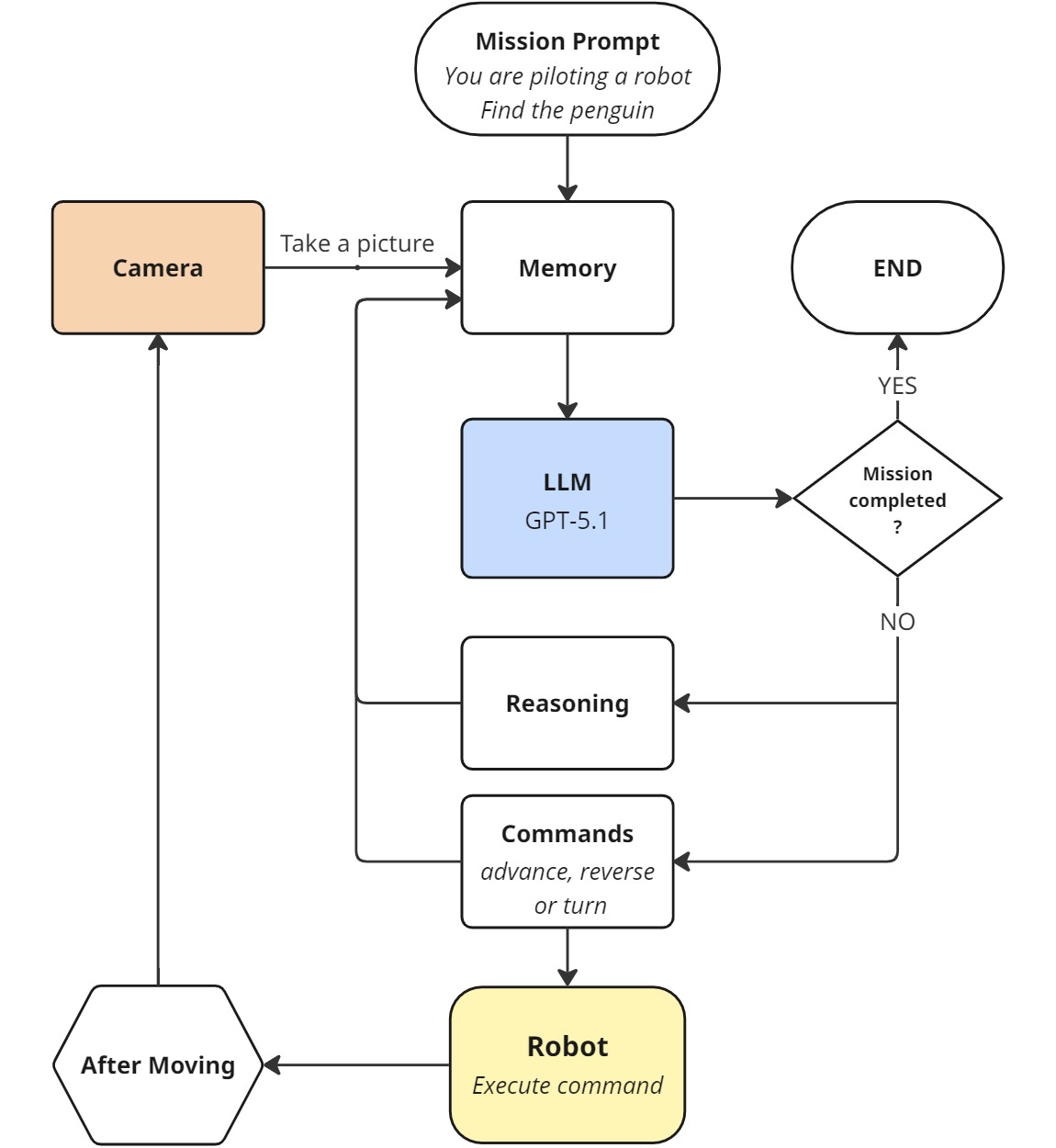}
\caption{System architecture showing the closed-loop interaction between the LLM (GPT-5.1), robot and camera.}
\label{fig:system_architecture}
\end{figure}

To evaluate the LLM as the control policy, we implemented a closed-loop architecture in which the model receives visual observations and produces motion commands iteratively. Figure~\ref{fig:system_architecture} illustrates the full system pipeline. The interaction protocol proceeds as follows:

1. The system prompt configures the LLM with the role: "You are piloting a small mobile robot..." A mission prompt defines the goal, such as locating a specific object (e.g., a toy penguin).
2. At each control cycle, the robot captures an image from its onboard camera. This image is transmitted to the LLM along with the current mission context.
3. The LLM analyzes the scene, generates a short textual reasoning trace, and outputs one command selected from a constrained action set: advance for $t$ seconds, reverse for $t$ seconds, or rotate by $\theta$ degrees.
4. The robot executes the commands. After motion completion, a new image is captured and sent back to the model, together with all previous images and reasoning traces, forming an accumulated short‑term memory. This expanded context is then used by the LLM to generate the next action, completing the perception–action loop.
5. The episode continues until the LLM explicitly declares "MISSION ACCOMPLISHED", which terminates the loop.

This architecture enables running fully embodied trials in real time, with the LLM operating as the high-level controller while the robot handles low-level motor execution.

\section{Results and Discussion}
In this exploratory study, we restricted our evaluation to GPT-5.1, one of the most advanced publicly available reasoning-capable multimodal language models at the time of experimentation, trained on text and images~\cite{openai2025gpt51systemcard}. 

We designed missions in which the robot was instructed to locate a specific target—a plush penguin—and physically touch it. When the robot was initially placed with the penguin already within its field of view, GPT-5.1 reliably identified the target and began issuing movement commands that generally resulted in forward progress toward the object. However, the model exhibited characteristic limitations in alignment control. While a human operator would typically rotate the robot until it faced the target head‑on and only then drive straight, the LLM tended to advance even when the penguin appeared off‑center, then adjust with small rotations, then advance again, in a spiral pattern.

To illustrate this behavior, we present the following run.

\subsection{Run 3 (2025-11-26)}
\begin{figure}[!t]
\centering
\includegraphics[width=1.0\columnwidth]{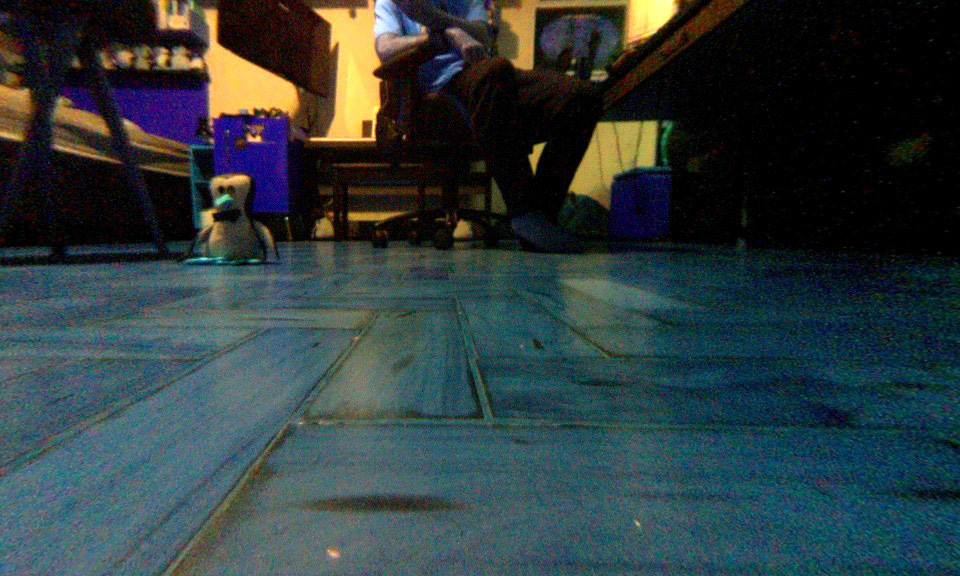}
\caption{Penguin is visible ahead-left.\protect\footnotemark}
\label{fig:run3_vision_012}
\end{figure}
\footnotetext{Colors look odd because the OpenCV library operates in BGR instead of RGB; this was corrected in the following runs.}
\begin{figure}[!t]
\centering
\includegraphics[width=1.0\columnwidth]{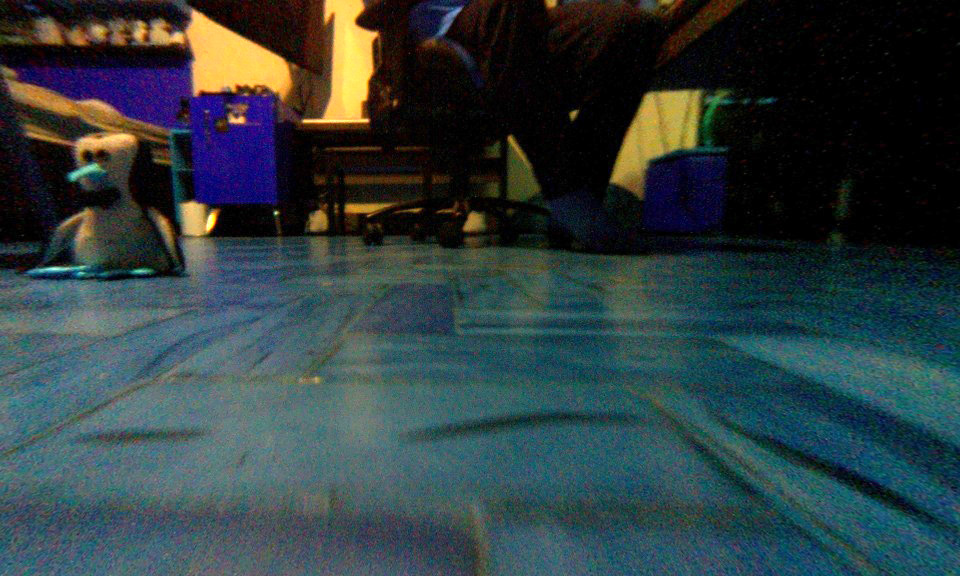}
\caption{Penguin is ahead-left, closer.}
\label{fig:run3_vision_013}
\end{figure}
\begin{figure}[!t]
\centering
\includegraphics[width=1.0\columnwidth]{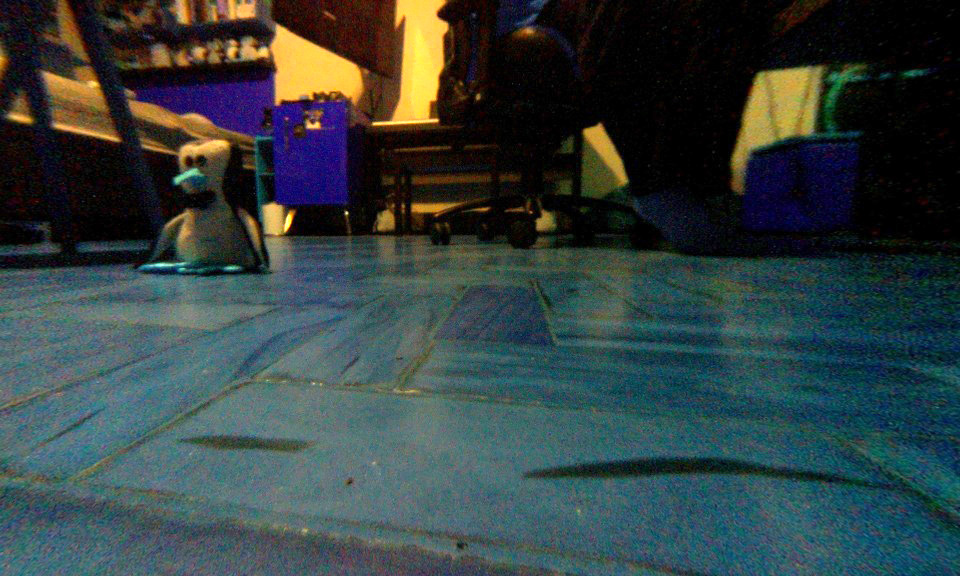}
\caption{After small rotation.}
\label{fig:run3_vision_014}
\end{figure}

Figure~\ref{fig:run3_vision_012}:
LLM reasoning: ``The penguin is clearly visible \textbf{ahead-left} with unobstructed floor between us, and we just need to come very close to it. I'll move \textbf{forward} a short distance to reduce the gap while keeping enough margin to stop before making contact and reassess our distance in the next frame.''\
Although the model correctly identified that the penguin was to the left, it chose to move forward first.
\
Command: \textit{advance 2 seconds (40 cm).}

Figure~\ref{fig:run3_vision_013}:
LLM reasoning: ``The penguin is \textbf{ahead-left} at close-to-moderate range with clear floor between us, but our heading is slightly to its right. I'll \textbf{turn a bit left} to better align with the penguin so we can approach it straight on while maintaining control over our stopping distance.''\
Command: \textit{turn left 10 degrees.}

Figure~\ref{fig:run3_vision_014}:
LLM reasoning: ``The penguin is directly \textbf{ahead-left} at fairly close range with clear floor in front of us. I'll \textbf{move forward} a short distance to come very close to it while minimizing risk of bumping into it. After this step we should be near enough to declare the mission complete.''\
The LLM rotated only slightly and again chose to advance rather than re-align.\
Command: \textit{advance 1 second (20 cm).}

This suboptimal alignment strategy did not occur in all trials—sometimes the model performed multiple corrective rotations—but it rarely aligned perfectly before advancing. Still, despite inefficiency, GPT‑5.1 typically managed to close distance consistently and complete the mission.

After initial tests, we increased task difficulty by placing the robot facing a wall, requiring the model to actively explore the environment before locating the penguin. GPT‑5.1 generally succeeded (4 out of 5) in maneuvering, even reversing when near obstacles (as confirmed by its reasoning traces), demonstrating a form of spatial awareness of its body in the environment.

\subsection{Run 2 (2025-11-27)}
\begin{figure}[!t]
\centering
\includegraphics[width=1.0\columnwidth]{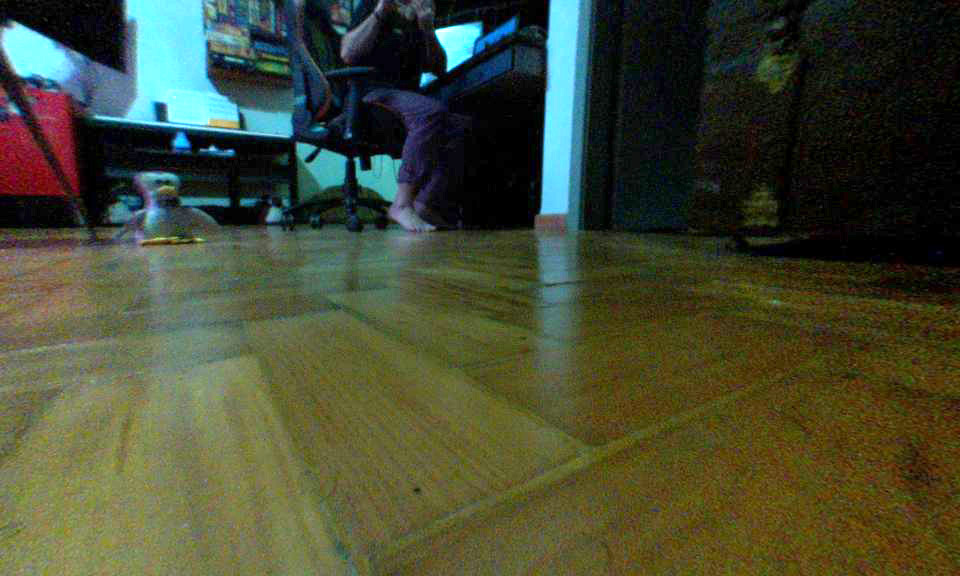}
\caption{Penguin is visible left of center.}
\label{fig:run2_vision_003}
\end{figure}
\begin{figure}[!t]
\centering
\includegraphics[width=1.0\columnwidth]{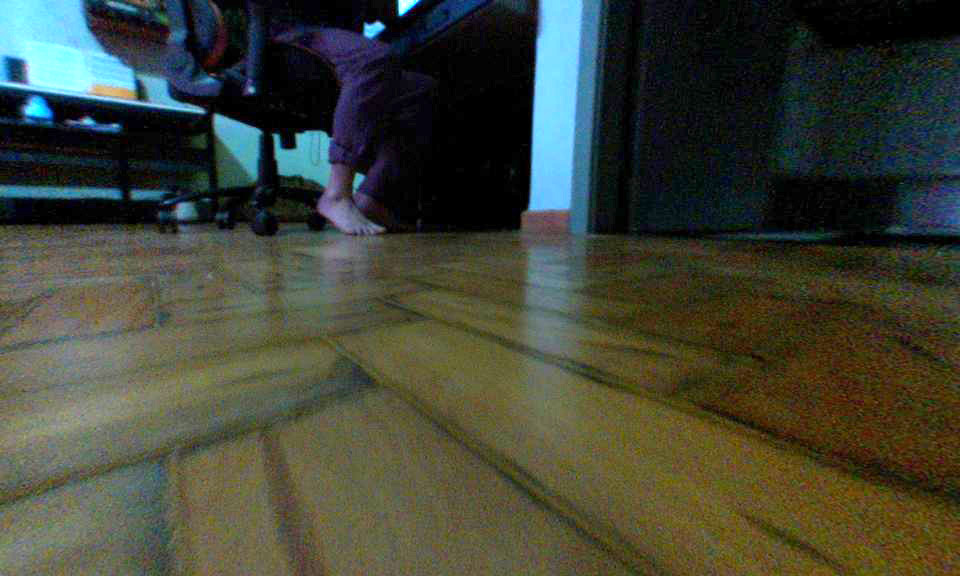}
\caption{Penguin left the frame. Model correctly infers its position.}
\label{fig:run2_vision_004}
\end{figure}
\begin{figure}[!t]
\centering
\includegraphics[width=1.0\columnwidth]{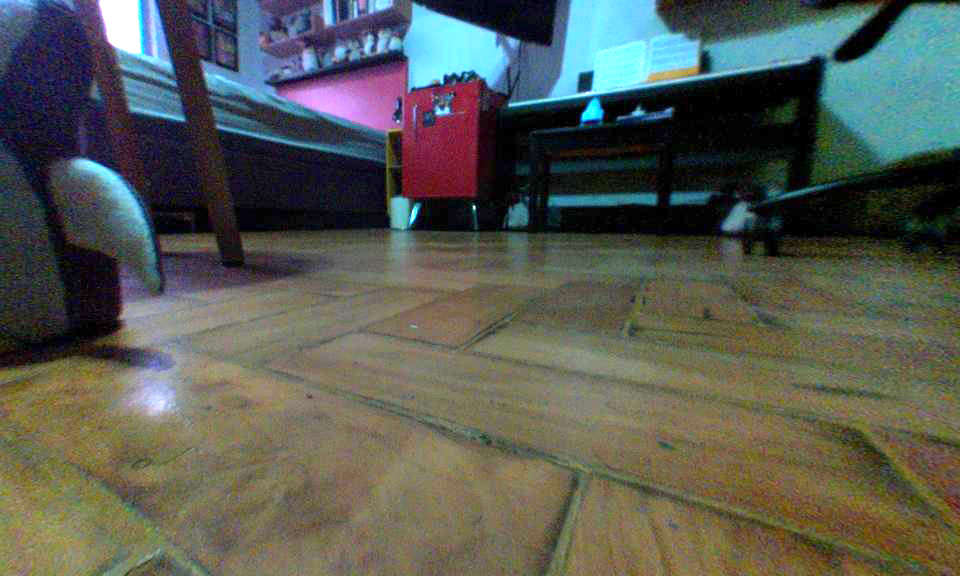}
\caption{Penguin visible on the left, after rotation.}
\label{fig:run2_vision_015}
\end{figure}
\begin{figure}[!t]
\centering
\includegraphics[width=1.0\columnwidth]{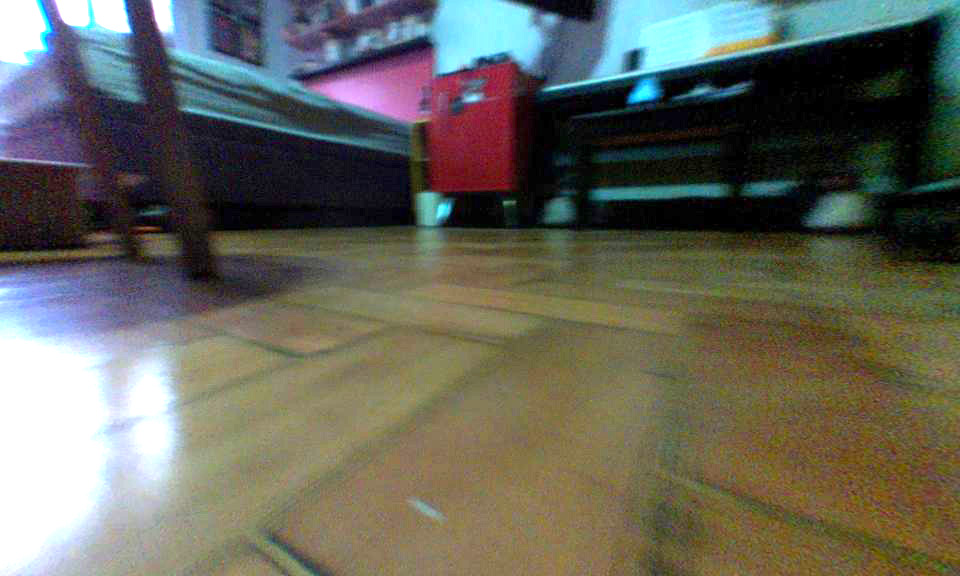}
\caption{Penguin out of view after small rotation.}
\label{fig:run2_vision_018}
\end{figure}
\begin{figure}[!t]
\centering
\includegraphics[width=1.0\columnwidth]{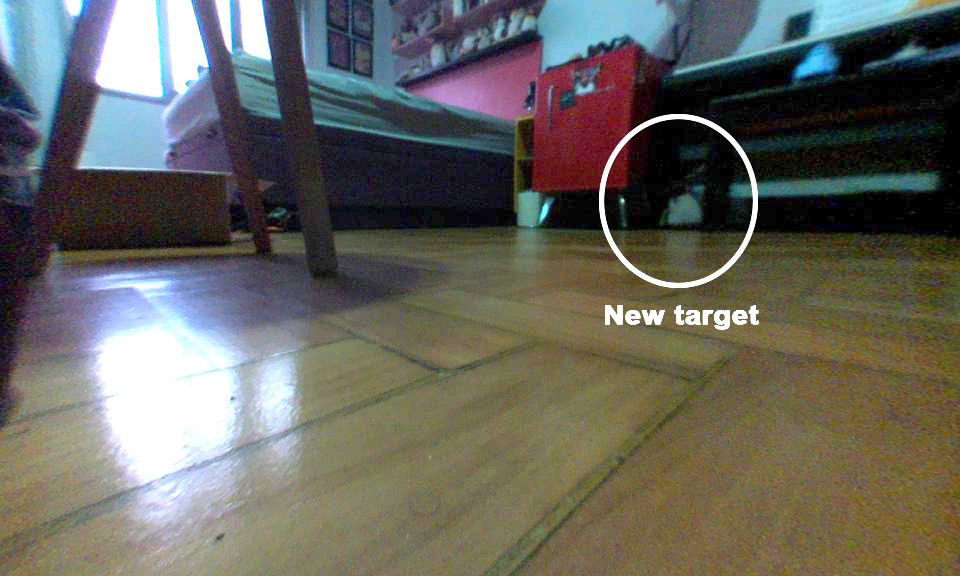}
\caption{Model misidentifies a distant penguin.}
\label{fig:run2_vision_019}
\end{figure}
\begin{figure}[!t]
\centering
\includegraphics[width=1.0\columnwidth]{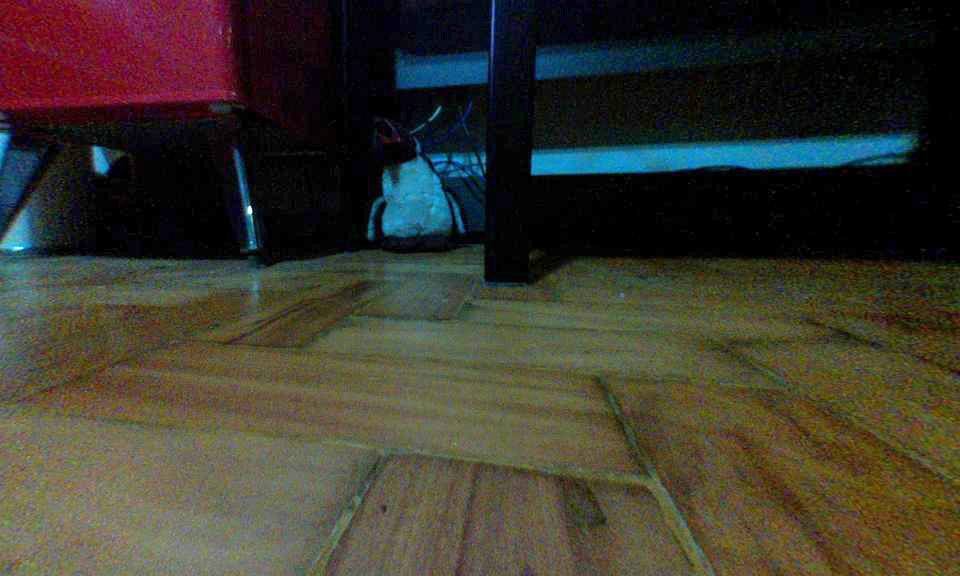}
\caption{Final frame of the sequence.}
\label{fig:run2_vision_035}
\end{figure}

Figure~\ref{fig:run2_vision_003}:
LLM reasoning: ``The penguin toy is visible slightly left of center at a medium distance on open floor. I'll drive straight toward it to close much of the gap while keeping it roughly in front of the robot. A 3‑second advance should move significantly closer without overshooting due to the wide‑angle distortion.''\
Command: \textit{advance 3 seconds (60 cm).}\
The robot advanced so far that the penguin left the field of view.

Figure~\ref{fig:run2_vision_004}:
This moment tests whether the model retains memory and spatial understanding: having previously seen the penguin on the left, it must infer that after advancing, the object should now be off-frame to its left.

LLM reasoning: ``The penguin is no longer visible, suggesting it is now \textbf{off to our left} after advancing. I'll \textbf{turn moderately left} to re-center the previous penguin location in the camera view while keeping clear of nearby furniture and feet.''\
Command: \textit{turn left 25 degrees.}\
This confirms an accurate spatial inference using only static images and accumulated context.

Figure~\ref{fig:run2_vision_015}:
After closing the distance using a spiral approach:
LLM reasoning: ``The penguin is clearly visible on the \textbf{left} and fairly close, with open floor ahead toward it. I'll \textbf{advance} a short distance to move closer and likely make contact without overshooting. This should make the penguin \textbf{grow significantly in the frame}; we can fine‑tune with very small moves afterward if it's not yet filling the view.''\
Command: \textit{advance 0.7 seconds (14 cm).}\
Given the penguin was far left, advancing was not optimal—it would not center the target nor increase its apparent size efficiently.

Figure~\ref{fig:run2_vision_018}:
LLM reasoning: ``The penguin, \textbf{previously at the extreme left}, is now out of view, meaning it is likely \textbf{just to our left or slightly behind} after the forward move. I'll \textbf{turn a bit further left} to sweep it back into the camera's field of view without changing our position much. Once it reappears and is closer to center, we can make a precise short advance to touch it and fill the frame.''\
Command: \textit{turn left 8 degrees.}\

Figure~\ref{fig:run2_vision_019}:
The rotation was too small; the penguin remained out of view.
At this point, the model mistakenly identified a distant penguin in the background.

LLM reasoning: ``The penguin is visible \textbf{ahead slightly right of center near the red cabinet}, with clear floor between us and it. I'll \textbf{move forward} a short distance to get closer while keeping it in view and avoiding the nearby furniture legs. This should noticeably increase its size in the frame so we can then fine‑tune our alignment and distance.''\
Command: \textit{advance 1 second (20 cm).}

Although GPT‑5.1 had previously inferred that the penguin was on the left, it now assumed that the distant plush on the right was the target, which is puzzling. Does the model lack the understanding that the original penguin could not have moved so far from its previous position, or did it simply abandon the earlier target and adopt a new one? Under the first interpretation, one might expect the LLM to notice, from its memory, that the far penguin was already present in the previous frame and therefore could not be the same object. However, this assumption may be unwarranted, because models do not perceive images in the same way humans do; it is possible that the distant penguin was never detected at all in the earlier, blurrier frame.

Ultimately, GPT‑5.1 pursued this new penguin and continued the mission accordingly (Figure~\ref{fig:run2_vision_035}).

\subsection{Run 6 (2025-11-27): Knock‑Down Task}
To further probe not only the model's spatial reasoning but also its understanding of physical interaction, we introduced a new mission: \textit{"find the penguin and knock it down with your head (camera)."}

\begin{figure}[!t]
\centering
\includegraphics[width=1.0\columnwidth]{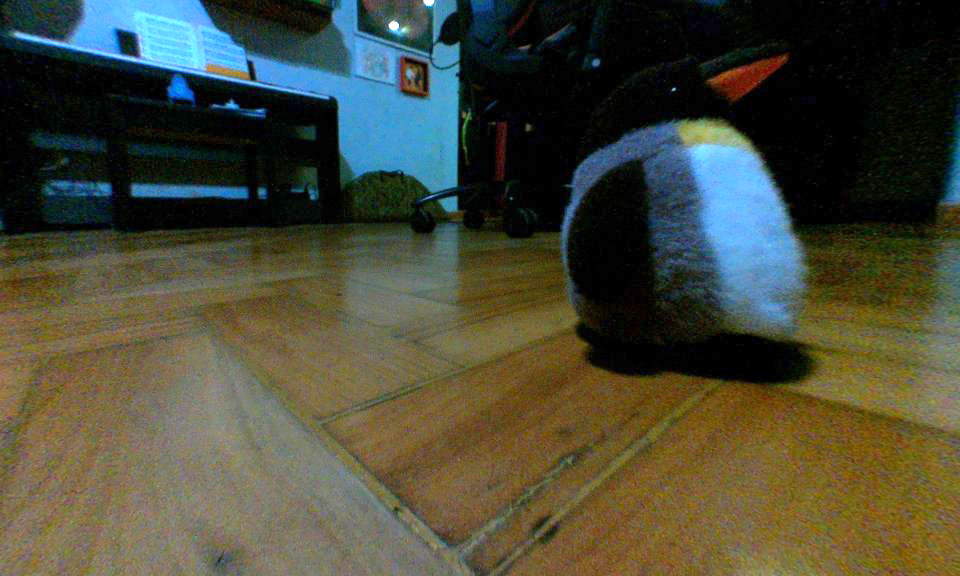}
\caption{Robot approaching the penguin before collision.}
\label{fig:run6_vision_014}
\end{figure}
\begin{figure}[!t]
\centering
\includegraphics[width=1.0\columnwidth]{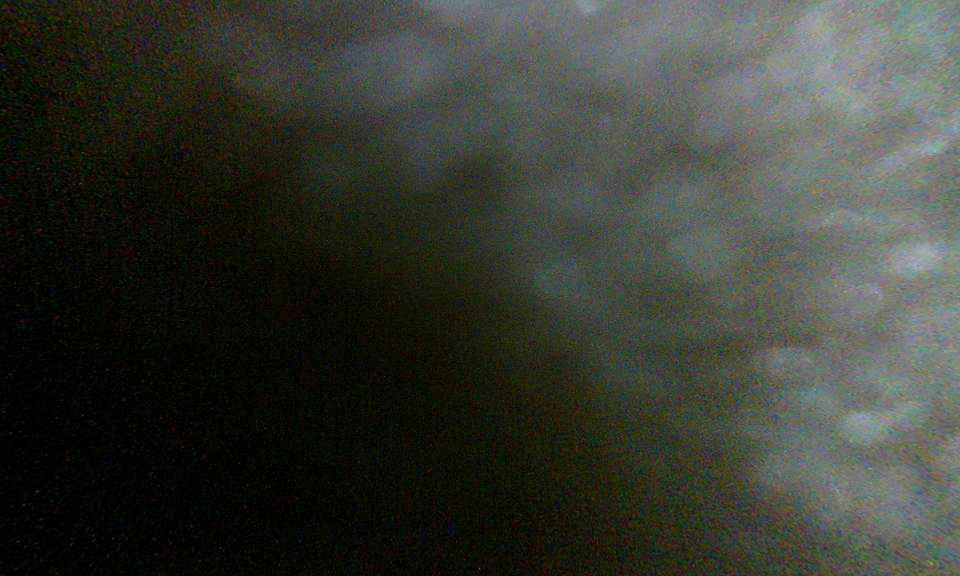}
\caption{After first impact, robot reverses to verify.}
\label{fig:run6_vision_025}
\end{figure}
\begin{figure}[!t]
\centering
\includegraphics[width=1.0\columnwidth]{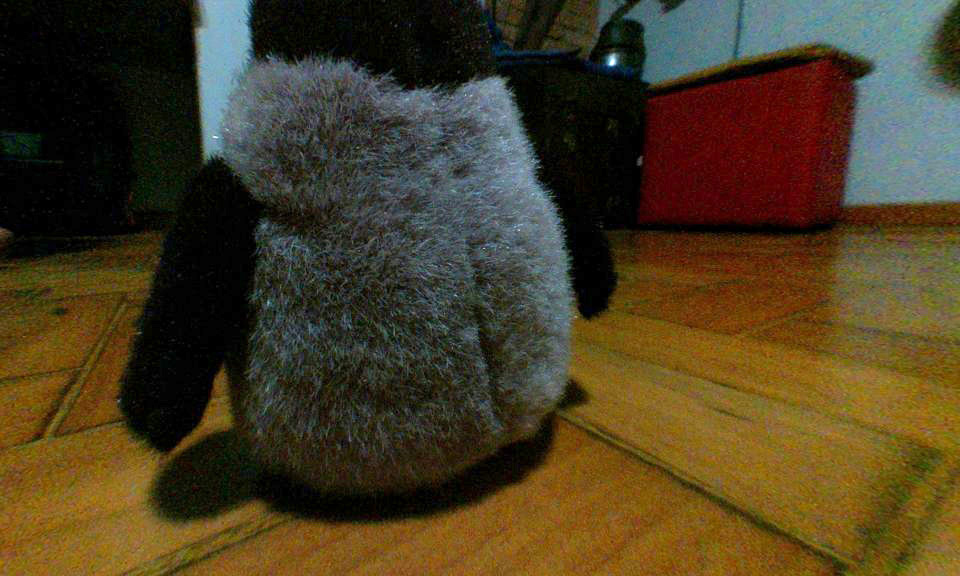}
\caption{Penguin still upright after first strike.}
\label{fig:run6_vision_026}
\end{figure}
\begin{figure}[!t]
\centering
\includegraphics[width=1.0\columnwidth]{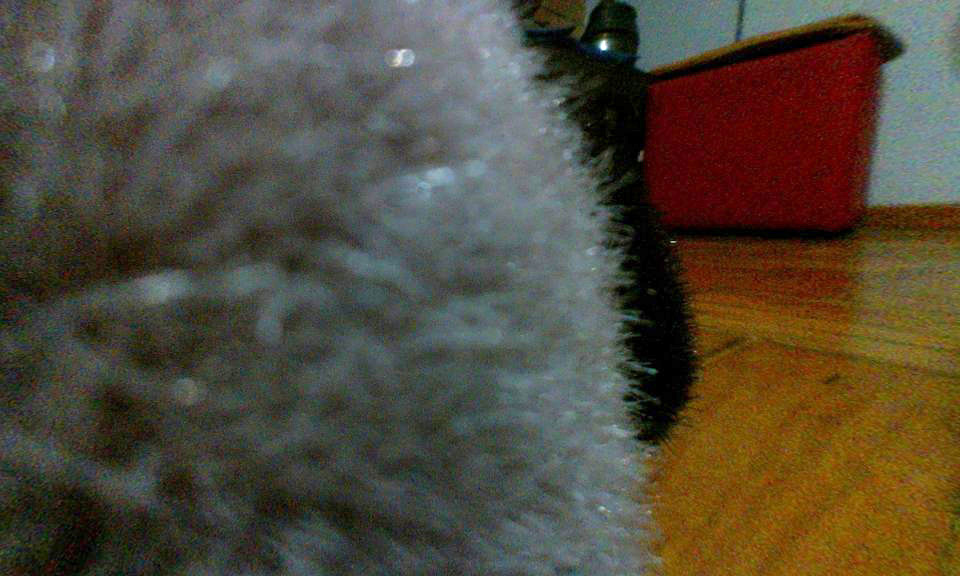}
\caption{Second strike.}
\label{fig:run6_vision_027}
\end{figure}
\begin{figure}[!t]
\centering
\includegraphics[width=1.0\columnwidth]{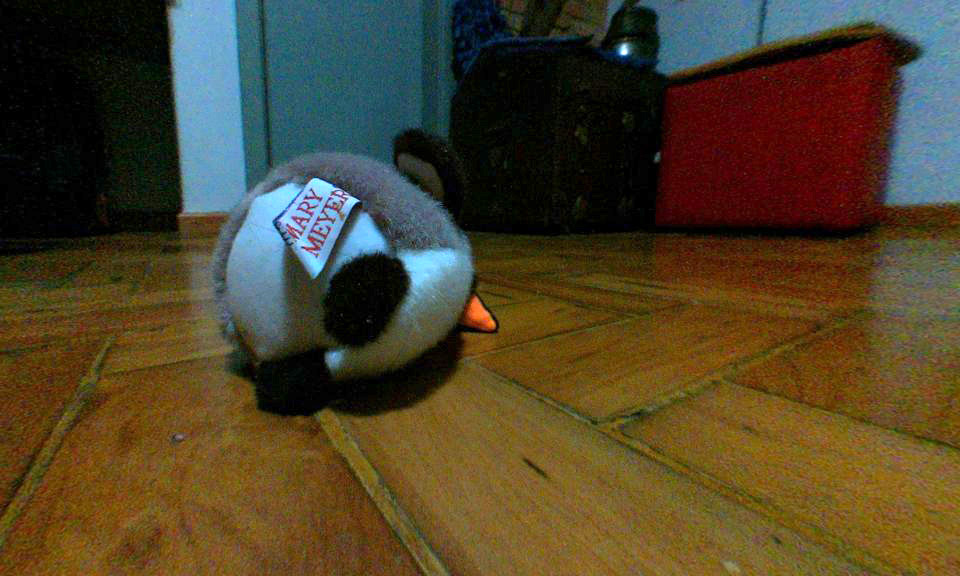}
\caption{Penguin toppled, mission complete.}
\label{fig:run6_vision_028}
\end{figure}

Figure~\ref{fig:run6_vision_014}: After locating the penguin and closing distance, the model successfully commanded the robot to collide with the target.

Figure~\ref{fig:run6_vision_025}: LLM reasoning: "We are pressed closely against the penguin's side, so it's hard to see whether it is upright or toppled. \textbf{Backing up} a short distance will create space so we can turn and visually confirm its orientation."\
Command: \textit{reverse 0.7 s (14 cm).}

This reasoning indicates that the model inferred physical contact and understood the need to reverse to regain visual confirmation—suggesting an emergent grasp of physical interaction despite no tactile sensors.

Figure~\ref{fig:run6_vision_026}: Upon seeing that the penguin was still upright, GPT‑5.1 initiated another forward strike.

Figures~\ref{fig:run6_vision_027} and~\ref{fig:run6_vision_028}: After the second impact, the robot again reversed to inspect the result. Detecting that the penguin was finally toppled, the model declared the mission complete.

\textbf{Aggregate results:} In the \textit{touch mission}, across five trials, the model failed only once, when it got stuck against a table leg (having no tactile sensors to feel the obstacle). In the \textit{knock‑down mission}, also across five trials, the model achieved two full successes, one partial outcome, and two failures. The partial case occurred when the robot located the penguin but did not manage to topple it, whereas the failures corresponded to runs in which the model did not find the penguin while navigating.

\section{Conclusion}
Despite being a short exploratory study using only a single model (GPT-5.1) and a limited scope of tasks, our experiments revealed striking indications of spatial and physical understanding emerging from a model trained solely on text and images. The robot’s camera feed consisted of low‑resolution (960×576), compressed (JPEG 85), and often noisy images, yet the model was able to navigate, orient itself, interpret spatial relations among objects, and even reason about physical interactions—all from static frames, one per action cycle.

GPT‑5.1 showed the ability to correctly infer the expected position of an object that left the camera's field of view as a consequence of its own forward motion. This required the model to maintain short‑term memory, track relative spatial changes, and perform causal inference—capacities not explicitly trained during pretraining. In the knock‑down task, the model exhibited a coherent action sequence: after colliding with the penguin, it reversed to visually confirm whether the object had toppled, repeated the strike when necessary, and then reassessed the result. Such behavior suggests a nontrivial internal representation of physical presence, contact, and proprioceptive awareness.

Nonetheless, limitations remain: the model often chose inefficient strategies such as spiral approaches, and occasionally misidentified distractor objects when the primary target left the frame—errors likely stemming from both incomplete spatial understanding and perceptual constraints.

Overall, despite these limitations, the experiments provide compelling evidence that GPT‑5.1 exhibits at least some degree of emergent spatial and physical understanding beyond its training regime, consistent with world-model-like behavior. Further, more controlled experiments are needed to characterize the boundaries, robustness, and generality of such capabilities in large multimodal language models.

\section*{Acknowledgment}

The authors thank the ``Futures'' network started by Max Nolan Chen, ``Caramelo Biônico'' and ``Fundo Patrimonial Amigos da Poli'' for providing resources and support.

\end{document}